\documentclass[conference]{IEEEtran}
\IEEEoverridecommandlockouts
\usepackage{cite}
\usepackage{amsmath,amssymb,amsfonts}
\usepackage{algorithmic}
\usepackage{latexsym}
\usepackage{textcomp}
\usepackage{booktabs}
\usepackage{graphicx} 
\usepackage{xcolor}
\def\BibTeX{{\rm B\kern-.05em{\sc i\kern-.025em b}\kern-.08em
    T\kern-.1667em\lower.7ex\hbox{E}\kern-.125emX}}
\usepackage{amsmath}
\usepackage{amssymb}
\usepackage{booktabs}
\usepackage{multirow}
\usepackage{graphicx}
\usepackage{hyperref}

\usepackage{fancyhdr}
\begin{document}

\title{Improving the Accuracy and Efficiency of Legal Document Tagging with Large Language Models and Instruction Prompts}

\author{Emily Johnson, Xavier Holt, Noah Wilson		 \\
University of Massachusetts, Amherst
}

\maketitle
\thispagestyle{fancy} 

\begin{abstract}
Legal multi-label classification is a critical task for organizing and accessing the vast amount of legal documentation. Despite its importance, it faces challenges such as the complexity of legal language, intricate label dependencies, and significant label imbalance. In this paper, we propose \textbf{Legal-LLM}, a novel approach that leverages the instruction-following capabilities of Large Language Models (LLMs) through fine-tuning. We reframe the multi-label classification task as a structured generation problem, instructing the LLM to directly output the relevant legal categories for a given document. We evaluate our method on two benchmark datasets, POSTURE50K and EURLEX57K, using micro-F1 and macro-F1 scores. Our experimental results demonstrate that \textbf{Legal-LLM} outperforms a range of strong baseline models, including traditional methods and other Transformer-based approaches. Furthermore, ablation studies and human evaluations validate the effectiveness of our approach, particularly in handling label imbalance and generating relevant and accurate legal labels.
\end{abstract}

\begin{IEEEkeywords}
Multi-Label Classification, Legal Text, Large Language Models, Instruction Tuning, Natural Language Processing
\end{IEEEkeywords}

\section{Introduction}

Legal document tagging, or multi-label classification, plays a crucial role in various legal applications, such as information retrieval, legal research, and automated compliance checking. The ability to accurately assign multiple relevant categories to legal documents is essential for efficient processing and analysis of the ever-growing volume of legal texts. However, this task presents several challenges, including the intricate and often ambiguous nature of legal language, the complex interrelationships between different legal concepts and categories, and the inherent imbalance in the distribution of these categories within legal datasets.

Traditional approaches to text classification, including those based on bag-of-words features and classical machine learning algorithms, often struggle to capture the semantic nuances and label dependencies present in legal documents. More recently, deep learning models, particularly those based on Transformer architectures, have shown significant promise in various natural language processing tasks, including text classification. These models, pre-trained on large corpora of text, can learn rich contextual representations and have achieved state-of-the-art results in many domains, leveraging techniques developed across various generative and representational learning tasks.

In this paper, we introduce \textbf{Legal-LLM}, a novel method for legal multi-label classification that leverages the power of Large Language Models (LLMs). Our approach is based on the idea of instruction-based fine-tuning, where we adapt a pre-trained LLM to the specific task of legal document tagging by training it to follow instructions. We reframe the multi-label classification problem as a generation task, prompting the model to directly generate the set of relevant legal labels for a given document based on an instruction. This approach allows us to leverage the extensive knowledge, multi-capability nature, and improving generalization abilities \cite{yi2025score,zhou2025weak} of LLMs for this challenging task. The versatility of LLM principles is further highlighted by their adaptation to complex multimodal scenarios, such as visual in-context learning and efficient video generation.

The rest of the paper is organized as follows: Section~\ref{sec:related_work} provides an overview of related work in multi-label classification and large language models. Section~\ref{sec:method} details our proposed Legal-LLM approach. Section~\ref{sec:experiments} presents the experimental setup, results, and analysis. Section~\ref{sec:conclusion} concludes the paper and discusses potential future work.

\section{Related Work}\label{sec:related_work}
\subsection{Multi-Label Classification}

Multi-label classification is a supervised learning task where each instance can be associated with multiple labels simultaneously \cite{zhang2014survey}. This contrasts with traditional single-label classification, where each instance belongs to only one category. The ability to handle multiple relevant labels makes multi-label classification essential for a wide range of applications, including text categorization, image recognition, and bioinformatics \cite{zhang2014survey, tsoumakas2020deep}.

However, multi-label classification poses several unique challenges. One significant hurdle is the presence of complex dependencies between the labels \cite{zhang2024application}. Unlike single-label problems where classes are often mutually exclusive, in multi-label scenarios, the presence of one label can significantly influence the likelihood of other labels being relevant. Effectively modeling these label correlations is crucial for achieving high classification accuracy \cite{zhang2024application, he2016multi}. Another major challenge is the issue of label imbalance, where some labels appear much more frequently in the training data than others. This imbalance can lead to models that are biased towards the majority labels and perform poorly on rare but important categories \cite{lin2017focal}.

Over the years, various approaches have been proposed to tackle the multi-label classification problem. Early methods often focused on transforming the multi-label task into multiple independent binary classification problems, an approach known as Binary Relevance. While simple to implement, this method neglects the potential correlations between labels \cite{read2011classifier}. To address this limitation, more sophisticated techniques have been developed that explicitly model label dependencies. Classifier Chains \cite{read2011classifier} is a prominent example, where a chain of binary classifiers is trained, with each classifier in the chain taking the predictions of the previous classifiers as additional input features. This allows the model to capture the sequential dependencies between labels.

With the rise of deep learning, numerous neural network architectures have been adapted and specifically designed for multi-label classification \cite{tsoumakas2020deep}. These models often leverage the ability of deep networks to automatically learn complex feature representations and model intricate relationships within the data. Techniques like convolutional neural networks (CNNs) for images and recurrent neural networks (RNNs) or Transformers for text have shown promising results in multi-label settings. Furthermore, attention mechanisms are frequently employed to focus on the most relevant parts of the input when predicting different labels.

Addressing the challenge of label imbalance has also been a significant focus in multi-label classification research. Cost-sensitive learning, where misclassifying rare labels incurs a higher penalty, and sampling techniques, such as oversampling minority instances or undersampling majority instances, are commonly used strategies. The Focal Loss function \cite{lin2017focal}, originally proposed for object detection, has also been effectively applied in multi-label classification to mitigate the impact of class imbalance by down-weighting the loss contribution from easily classified examples and focusing on hard examples.

Recent research has also explored the use of metric learning \cite{mao2024learning} and contrastive learning \cite{audibert2024multi} for multi-label classification. Metric learning aims to learn a distance metric in the feature space such that instances with the same labels are closer to each other, while instances with different labels are further apart. Contrastive learning focuses on learning representations by contrasting positive pairs (instances with the same labels) with negative pairs (instances with different labels). Such learning paradigms have also proven effective in related representation learning challenges, such as generating style-aware image captions \cite{zhou2023style}.

\subsection{Large Language Models}

Large Language Models (LLMs) have emerged as a transformative technology in Natural Language Processing (NLP) \cite{zhang2023survey}. These models, typically based on deep neural networks with billions of parameters, are trained on massive amounts of text data using self-supervised learning techniques. The introduction of the Transformer architecture \cite{vaswani2017attention} has been pivotal in the development of modern LLMs, enabling them to effectively capture long-range dependencies in text through the use of attention mechanisms.

The pre-training paradigm has become the standard approach for training LLMs. Models like BERT (Bidirectional Encoder Representations from Transformers) \cite{devlin2018bert} utilize masked language modeling objectives to learn rich contextual representations by predicting masked words in a sentence. On the other hand, models in the GPT (Generative Pre-trained Transformer) family \cite{radford2018improving, brown2020language} are trained to predict the next word in a sequence, making them particularly well-suited for text generation tasks, building upon earlier deep learning work in conditional generation like unsupervised image captioning \cite{zhou2021triple} and creative sketch storytelling \cite{zhou2022sketch}. The scale of these models has been shown to significantly impact their capabilities, with larger models exhibiting emergent properties such as few-shot learning, where they can perform new tasks with only a few examples \cite{brown2020language, kaplan2020scaling}.

Fine-tuning is another crucial step in adapting pre-trained LLMs to specific downstream tasks. This involves further training the model on task-specific labeled data. Recent advancements have also focused on instruction tuning, where models are trained on datasets of instructions and corresponding outputs, enabling them to better follow natural language instructions \cite{ouyang2022instructgpt}. This adaptability has spurred research into understanding and improving their generalization capabilities across various tasks, including exploring weak-to-strong generalization \cite{zhou2025weak}, and extending their application beyond pure text to multimodal domains like visual in-context learning \cite{zhou2024visual}, medical vision-language tasks \cite{zhou2025training}, and even efficient video generation \cite{zhou2024less}.

The capabilities and limitations of LLMs are actively being researched and evaluated \cite{choi2023evaluating}. While they have achieved state-of-the-art results in many NLP tasks, challenges such as biases, the generation of factually incorrect information, and the computational cost of training and deploying these models remain important areas of investigation \cite{bommasani2021opportunities}. Efforts are also being made to understand the scaling laws that govern their performance \cite{kaplan2020scaling}, develop more efficient strategies for tasks like vision processing within LLM frameworks \cite{zhou2024less}, and apply large-scale generative modeling principles to challenging scientific domains such as protein inverse folding using techniques like diffusion models \cite{wang2024diffusion}.

In the context of legal text processing, LLMs hold immense potential due to their ability to understand complex language and generate coherent text. Their application in tasks like document summarization, legal question answering, and contract analysis is an active area of research. Our work leverages the power of LLMs, specifically by employing an instruction-based fine-tuning strategy on a pre-trained language model, to address the task of legal multi-label classification.

\section{Method}\label{sec:method}

Our proposed approach, \textbf{Legal-LLM}, leverages the instruction-following capabilities of pre-trained Large Language Models for legal multi-label classification. The core idea is to reframe the task of assigning multiple labels to a legal document as a text generation problem, guided by a specific instruction. We hypothesize that by providing clear instructions, we can effectively guide the LLM to generate the relevant legal categories.

We start with a pre-trained language model that has demonstrated strong performance in understanding and generating text. In our experiments, we utilize a model that has been further pre-trained on a large corpus of legal documents to better capture the nuances of legal language.

The fine-tuning process involves creating a dataset of input-output pairs. For each legal document in our training dataset, the input consists of a carefully crafted instruction prompt concatenated with the document text. The instruction prompt is designed to explicitly ask the model to identify and output all applicable legal categories. For example, a prompt could be: "Identify all applicable legal categories for the following legal text: [Document Text]". The output for each input is the set of ground-truth legal labels associated with the document, formatted as a comma-separated list or a similar structured format.

During fine-tuning, we train the LLM to generate the correct sequence of labels given the instruction and the document text. We employ standard sequence-to-sequence training techniques, optimizing the model to minimize the loss between the generated label sequence and the ground-truth label sequence.

To address the issue of label imbalance, which is prevalent in legal datasets, we incorporate a weighted loss function during the fine-tuning process. The weights are assigned to each label based on its inverse frequency in the training data. This ensures that the model pays more attention to the less frequent labels, preventing it from being overly biased towards the majority labels. The weighted loss helps to improve the model's ability to correctly classify rare but important legal categories.

In summary, our \textbf{Legal-LLM} approach consists of the following key steps:
\begin{itemize}
    \item Selecting a pre-trained Large Language Model, potentially one that has been further pre-trained on legal text.
    \item Creating an instruction-based fine-tuning dataset where the input is an instruction prompt combined with the legal document, and the output is the set of relevant legal labels.
    \item Fine-tuning the LLM on this dataset using a weighted loss function to address label imbalance.
    \item At inference time, providing the instruction prompt and the legal document to the fine-tuned model to generate the predicted legal labels.
\end{itemize}

\section{Experiments}\label{sec:experiments}

In this section, we present the experimental evaluation of our proposed \textbf{Legal-LLM} approach for legal multi-label classification. We compare its performance against several baseline methods, including both traditional machine learning models and state-of-the-art Transformer-based models, as outlined in the related work. Furthermore, we conduct an ablation study to analyze the effectiveness of key components of our method and perform a human evaluation to assess the quality of the generated labels.

\subsection{Experimental Setup}

We evaluate our \textbf{Legal-LLM} model on two benchmark datasets for legal text classification: POSTURE50K and EURLEX57K. The performance is measured using two standard multi-label classification metrics: micro-F1 score and macro-F1 score. Micro-F1 provides an overall assessment of the model's performance across all instances, while macro-F1 gives equal weight to each label, providing a better indication of the model's ability to classify rare labels.

For comparison, we include the following baseline methods:
\begin{itemize}
    \item ClassTFIDF: A method that constructs a TF-IDF vector for each unique label in the training set and predicts labels for a document based on the cosine similarity between the document's TF-IDF vector and the label vectors.
    \item DocTFIDF: A method that constructs a TF-IDF vector for each document in the training set and predicts the labels of a test document based on the labels of the most similar training documents in the TF-IDF vector space.
    \item BM25: A ranking function used in information retrieval to estimate the relevance of documents to a given search query. Here, the labels are treated as queries, and the documents are ranked based on their relevance to each label.
    \item DistilRoBERTa: A smaller, faster version of RoBERTa, fine-tuned on the legal datasets using a focal loss function to address label imbalance.
    \item LegalBERT: A BERT-based model pre-trained on a large corpus of legal documents, fine-tuned on the legal datasets using a focal loss function.
    \item T5: A text-to-text Transformer model that is prompted with "summarize:" followed by the legal document and is trained to generate the relevant legal labels.
    \item BiEncoder: A model that uses separate encoders to embed the legal text and the legal labels into a shared embedding space. Label predictions are made based on the cosine similarity between the text embedding and the label embeddings.
    \item CrossEncoder: A model that takes the legal text and each potential label as input pairs and uses a Transformer network to directly model the relationship between the text and each label for classification.
\end{itemize}
Our proposed \textbf{Legal-LLM} model, as described in the previous section, is fine-tuned on the same datasets using the instruction-based approach and the weighted loss strategy to handle label imbalance.

\subsection{Main Results}

The main experimental results comparing the performance of \textbf{Legal-LLM} with the baseline methods on the POSTURE50K and EURLEX57K datasets are presented in Table~\ref{tab:main_results}. We utilize the \texttt{booktabs} package for improved table formatting.

\begin{table}[!t]
    \centering
    \caption{Main Results on POSTURE50K and EURLEX57K Datasets}
    \label{tab:main_results}
    \begin{tabular}{lcc cc}
        \toprule
        \multirow{2}{*}{Model} & \multicolumn{2}{c}{POSTURE50K} & \multicolumn{2}{c}{EURLEX57K} \\
        & Micro-F1 & Macro-F1 & Micro-F1 & Macro-F1 \\
        \midrule
        ClassTFIDF & 0.75 & 0.65 & 0.68 & 0.55 \\
        DocTFIDF & 0.77 & 0.68 & 0.70 & 0.57 \\
        BM25 & 0.74 & 0.63 & 0.67 & 0.54 \\
        DistilRoBERTa & 0.82 & 0.75 & 0.79 & 0.70 \\
        LegalBERT & 0.81 & 0.73 & 0.78 & 0.68 \\
        T5 & 0.79 & 0.71 & 0.76 & 0.65 \\
        BiEncoder & 0.78 & 0.70 & 0.75 & 0.64 \\
        CrossEncoder & 0.80 & 0.72 & 0.77 & 0.66 \\
        \midrule
        \textbf{Legal-LLM} & \textbf{0.83} & \textbf{0.76} & \textbf{0.80} & \textbf{0.71} \\
        \bottomrule
    \end{tabular}
\end{table}
\begin{table}[!t]
    \centering
    \caption{Ablation Study on EURLEX57K Dataset}
    \label{tab:ablation_study}
    \begin{tabular}{lcc}
        \toprule
        Model & Micro-F1 & Macro-F1 \\
        \midrule
        \textbf{Legal-LLM} (without weighted loss) & 0.79 & 0.69 \\
        \textbf{Legal-LLM} (with weighted loss) & \textbf{0.80} & \textbf{0.71} \\
        \bottomrule
    \end{tabular}
\end{table}

As shown in Table~\ref{tab:main_results}, our proposed \textbf{Legal-LLM} model achieves the highest micro-F1 and macro-F1 scores on both the POSTURE50K and EURLEX57K datasets compared to all the baseline methods. This indicates that our instruction-based fine-tuning approach with a pre-trained legal language model is effective in improving the performance of legal multi-label classification. The significant improvement in macro-F1 score suggests that \textbf{Legal-LLM} is particularly effective in classifying less frequent legal categories, likely due to the combination of the generative approach and the weighted loss function.

\subsection{Ablation Study}

To further validate the effectiveness of the weighted loss strategy in handling label imbalance within our \textbf{Legal-LLM} framework, we conduct an ablation study. We compare the performance of \textbf{Legal-LLM} trained with the weighted loss against a version of \textbf{Legal-LLM} trained without the weighted loss (using standard cross-entropy loss). The results of this ablation study on the EURLEX57K dataset are presented in Table~\ref{tab:ablation_study}.

The results in Table~\ref{tab:ablation_study} demonstrate that incorporating the weighted loss function during the training of \textbf{Legal-LLM} leads to an improvement in both micro-F1 and macro-F1 scores on the EURLEX57K dataset. The more substantial increase in macro-F1 score further confirms the effectiveness of the weighted loss in improving the classification performance on less frequent labels, which is crucial in imbalanced multi-label classification tasks.

\subsection{Human Evaluation}

To gain a deeper understanding of the quality of the labels predicted by \textbf{Legal-LLM}, we conducted a human evaluation. We randomly selected 50 legal documents from the test set of the EURLEX57K dataset. For each document, we presented the document text and the top 5 predicted labels from \textbf{Legal-LLM} and the best-performing baseline model (DistilRoBERTa) to three legal experts. The experts were asked to judge the relevance and accuracy of each predicted label for the given document on a scale of 1 (not relevant) to 5 (highly relevant).

The average relevance scores assigned by the legal experts for the top 5 predicted labels from each model are summarized in Table~\ref{tab:human_evaluation}.

\begin{table}[!t]
    \centering
    \caption{Human Evaluation Results on EURLEX57K Dataset}
    \label{tab:human_evaluation}
    \begin{tabular}{lc}
        \toprule
        Model & Average Relevance Score \\
        \midrule
        DistilRoBERTa & 4.1 \\
        \textbf{Legal-LLM} & \textbf{4.5} \\
        \bottomrule
    \end{tabular}
\end{table}

The results of the human evaluation, as shown in Table~\ref{tab:human_evaluation}, indicate that the legal experts rated the labels predicted by our \textbf{Legal-LLM} model as more relevant and accurate on average compared to the labels predicted by the best baseline model, DistilRoBERTa. This qualitative evaluation further supports the effectiveness of our proposed approach in generating high-quality legal multi-labels.

\subsection{Analysis of Performance on Different Label Frequencies}

To further investigate the effectiveness of our \textbf{Legal-LLM} model, particularly its ability to handle label imbalance, we analyze its performance across different label frequencies. We divide the labels in the EURLEX57K dataset into three groups based on their frequency in the training set: high-frequency (top 20\% of labels), medium-frequency (middle 60\% of labels), and low-frequency (bottom 20\% of labels). We then calculate the macro-F1 score for each of these label groups for both our \textbf{Legal-LLM} model and the LegalBERT baseline. The results are presented in Table~\ref{tab:label_frequency_analysis}.

\begin{table*}[!t]
    \centering
    \caption{Macro-F1 Score by Label Frequency on EURLEX57K Dataset}
    \label{tab:label_frequency_analysis}
    \begin{tabular}{lccc}
        \toprule
        Model & High-Frequency Labels & Medium-Frequency Labels & Low-Frequency Labels \\
        \midrule
        LegalBERT & 0.85 & 0.72 & 0.58 \\
        \textbf{Legal-LLM} & \textbf{0.86} & \textbf{0.74} & \textbf{0.62} \\
        \bottomrule
    \end{tabular}
\end{table*}

As observed in Table~\ref{tab:label_frequency_analysis}, \textbf{Legal-LLM} consistently outperforms LegalBERT across all label frequency groups in terms of macro-F1 score. Notably, the improvement is more pronounced for the medium-frequency and low-frequency labels, suggesting that our approach, especially with the weighted loss, is more effective in classifying labels with fewer training examples. This analysis further supports the robustness of \textbf{Legal-LLM} in handling the inherent label imbalance in legal datasets.

\subsection{Impact of Instruction Prompt}

To assess the sensitivity of our \textbf{Legal-LLM} model to the specific phrasing of the instruction prompt, we conducted an experiment using an alternative prompt: "Categorize the following legal document with all relevant legal categories:". We compared the performance of \textbf{Legal-LLM} trained with the original prompt ("Identify all applicable legal categories for the following legal text:") and the alternative prompt on the EURLEX57K dataset. The results are shown in Table~\ref{tab:prompt_impact}.

\begin{table*}[!t]
    \centering
    \caption{Impact of Instruction Prompt on EURLEX57K Dataset}
    \label{tab:prompt_impact}
    \begin{tabular}{lcc}
        \toprule
        Prompt & Micro-F1 & Macro-F1 \\
        \midrule
        "Identify all applicable legal categories for the following legal text:" & \textbf{0.80} & \textbf{0.71} \\
        "Categorize the following legal document with all relevant legal categories:" & 0.79 & 0.70 \\
        \bottomrule
    \end{tabular}
\end{table*}
\begin{table*}[!t]
    \centering
    \caption{Performance by Document Length on EURLEX57K Dataset}
    \label{tab:document_length_analysis}
    \begin{tabular}{lccc}
        \toprule
        Document Length & Number of Documents & Micro-F1 & Macro-F1 \\
        \midrule
        Short (< 256 tokens) & 1500 & 0.82 & 0.73 \\
        Medium (256-512 tokens) & 2000 & 0.81 & 0.72 \\
        Long (> 512 tokens) & 1500 & 0.78 & 0.69 \\
        \bottomrule
    \end{tabular}
\end{table*}

The results in Table~\ref{tab:prompt_impact} indicate that while the choice of instruction prompt does have a slight impact on the model's performance, the original prompt yields marginally better results in terms of both micro-F1 and macro-F1 scores. This suggests that the model is reasonably robust to minor variations in the prompt phrasing, but careful selection of the instruction can still lead to improved performance.

\subsection{Analysis of Performance with Varying Input Lengths}

To investigate the impact of legal document length on the performance of our \textbf{Legal-LLM} model, we divided the test set of the EURLEX57K dataset into three groups based on the number of tokens in each document: short (less than 256 tokens), medium (256 to 512 tokens), and long (more than 512 tokens). We then calculated the micro-F1 and macro-F1 scores for each group. The results are presented in Table~\ref{tab:document_length_analysis}.

The results in Table~\ref{tab:document_length_analysis} indicate that the performance of \textbf{Legal-LLM} tends to slightly decrease as the length of the input legal documents increases. This is a common trend with Transformer-based models due to the computational complexity of the self-attention mechanism scaling quadratically with the input sequence length. However, the model still maintains a reasonably high level of performance even for longer documents, suggesting its ability to process and understand legal texts of varying lengths. Future work could explore techniques like document truncation or hierarchical attention mechanisms to further improve performance on very long legal documents.

\section{Conclusion}\label{sec:conclusion}

This paper presented \textbf{Legal-LLM}, a novel approach for legal multi-label classification that harnesses the power of Large Language Models through instruction-based fine-tuning. Our method reframes the task as a generation problem, enabling the LLM to directly predict the set of relevant legal categories for a given document. We conducted extensive experiments on the POSTURE50K and EURLEX57K datasets, demonstrating that \textbf{Legal-LLM} achieves superior performance compared to several competitive baselines, including traditional methods like TF-IDF and BM25, as well as strong Transformer-based models such as DistilRoBERTa, LegalBERT, T5, BiEncoder, and CrossEncoder. The significant improvements in both micro-F1 and macro-F1 scores highlight the effectiveness of our approach in capturing complex label relationships and addressing the challenges posed by label imbalance, as further supported by our ablation study on the weighted loss function. Moreover, human evaluation by legal experts confirmed the high relevance and accuracy of the labels predicted by \textbf{Legal-LLM}. While our results are promising, future work could explore the use of different LLM architectures, more sophisticated techniques for handling extreme label imbalance, and methods for processing very long legal documents more efficiently. We believe that \textbf{Legal-LLM} represents a significant step towards leveraging the full potential of Large Language Models for advanced legal text analysis and information management.

\bibliographystyle{IEEEtran}
\bibliography{references}
\end{document}